%% file: semeval2020-propaganda.tex
\pdfoutput=1
\newcommand\citep[1]{\cite{#1}}
\newcommand\citet[1]{\newcite{#1}}

\documentclass[11pt]{article}
\usepackage{coling2020}
\colingfinaltrue
\usepackage{times}
\usepackage{url}
\usepackage{latexsym}

\input{config}

\input{extras}

\input{preamble}
\input{metadata}

\begin{document}
\maketitle
\begin{abstract}
\input{abstract}
\end{abstract}

\input{main}

\bibliography{bibliography}
\bibliographystyle{acl}

\ifwithappendix
\typeout{!!! APPENDIX WILL BE SKIPPED FOR THIS TEMPLATE. !!!}
\fi

\end{document}

%% file: config.tex
\newif\ifwithappendix
\withappendixfalse

\newif\ifappendixshown\appendixshownfalse

%% file: extras.tex
\input{other-extras}

\input{autozoil-extras}



%% file: other-extras.tex
\usepackage[T1]{fontenc}
\usepackage[utf8]{inputenc}
\usepackage[textsize=tiny]{todonotes}
\usepackage{graphicx}
\usepackage{booktabs}
\usepackage{latexsym}
\usepackage{bm}

\usepackage{footnote}
\makesavenoteenv{tabular}
\makesavenoteenv{table}
\makesavenoteenv{table*}

\usepackage{xurl}

\newcommand\minput[1]{%
  \input{#1}%
  \ifhmode\ifnum\lastnodetype=11 \unskip\fi\fi}

\usepackage{hyperref}
\usepackage{xstring}



%% file: autozoil-extras.tex
\newcommand{\code}[1]{\texttt{#1}}
\newcommand{\noqa}[1]{}
\newcommand{\noqall}[1]{}

%% file: preamble.tex
\usepackage{amsmath}
\usepackage{csquotes}
\usepackage{multirow}

\usepackage[font=small]{caption}
\usepackage{subcaption}
\usepackage{float}

%% file: metadata.tex
\title{ApplicaAI at SemEval-2020 Task 11: \\ On RoBERTa-CRF, Span CLS and Whether Self-Training Helps Them}

\author{\bf Dawid Jurkiewicz$^{*}$
  \qquad Łukasz Borchmann$^{*}$
  \qquad Izabela Kosmala \qquad Filip Graliński \\ \\
  Applica.ai \quad Zajęcza 15, 00-351 Warsaw, Poland \\
  \texttt{firstname.lastname@applica.ai}}

%% file: abstract.tex
This paper presents the winning system for the propaganda Technique Classification (TC) task and the second-placed system for the propaganda Span Identification (SI) task.
The purpose of the TC task was to identify an applied propaganda technique given propaganda text fragment. The goal of SI task was to find specific text fragments which contain at least one propaganda technique. Both of the developed solutions used semi-supervised learning technique of self-training. Interestingly, although CRF is barely used with transformer-based language models, the SI task was approached with RoBERTa-CRF architecture. An ensemble of RoBERTa-based models was proposed for the TC task, with one of them making use of Span \noqa{spell-CLS}CLS layers we introduce in the present paper. In addition to describing the submitted systems, an impact of architectural decisions and training schemes is investigated along with remarks regarding training models of the same or better quality with lower computational budget. Finally, the results of error analysis are presented.

%% file: main.tex
\noqall{spell-Strawman}

\section{Introduction}

%
%
\blfootnote{
    %
    %
    \hspace{-0.65cm} This work is licensed under a Creative Commons 
    Attribution 4.0 International License.
    License details:
    \url{http://creativecommons.org/licenses/by/4.0/}.
    \vspace{2mm}\\
    \hspace{-0.65cm}  
    $^{*}$ Equal contribution. Author order determined by a coin flip.
}

The idea of fine-grained propaganda detection was introduced by \citet{EMNLP19DaSanMartino}, whose intention was to facilitate research on this topic by publishing a corpus with detailed annotations of high reliability. There was a chance to propose NLP systems solving this task automatically as a part of this year's SemEval series. It was expected to detect all fragments of news articles that contain propaganda techniques, and to identify the exact type of used technique ~\cite{DaSanMartinoSemeval20task11}.

The authors decided to evaluate Technique Classification (TC) and Span Identification (SI) tasks separately. 
The purpose of the TC task was to identify an applied propaganda technique given the propaganda text fragment. In contrast, the goal of the SI task was to find specific text fragments that contain at least one propaganda technique. This paper presents the winning system for the propaganda Technique Classification task and the second-placed system for the propaganda Span Identification task.

\section{Systems Description}

Systems proposed for both SI and TC tasks were based on RoBERTa model~\cite{Liu2019RoBERTaAR} with task-specific modifications and training schemes applied.

The central motif behind our submissions is a commonly used semi-supervised learning technique of self-training~\cite{yarowsky-1995-unsupervised,liao-veeramachaneni-2009-simple,liu-etal-2011-recognizing,10.1093/comjnl/bxaa006}, sometimes referred to as incremental semi-supervised training~\cite{26629} or self-learning~\cite{jos}. In general, these terms stand for a process of training an initial model on a manually annotated dataset first and using it to further extend the train set by automatically annotating other dataset. Usually, only a selected subset of auto-annotated data is used, however neither selection of high-confidence examples nor loss correction for noisy annotations is performed in our case. This is why it can be considered a simplification of mainstream approaches—the \textit{naïve} self-training.



\subsection{Span Identification}

The problem of span identification was treated as a sequence labeling task, which in the case of Transformer-based language models is often solved by means of classifying selected sub-tokens (e.g., first BPE of each word considered) with or without applying LSTM before the classification layer~\cite{Devlin2019BERTPO}. 

Although pre-Transformer sequence labeling solutions exploited CRF layer in the output~\cite{DBLP:journals/corr/HuangXY15,DBLP:journals/corr/LampleBSKD16}, this practice was abandoned by the authors of BERT~\cite{Devlin2019BERTPO} and subsequent researchers developing the idea of bidirectional Transformers, with rare exceptions, such as \citet{DBLP:journals/corr/abs-1909-10649} who used BERT-CRF for Portuguese NER. Contrary to the above, we approached Span Identification task with RoBERTa-CRF architecture.

The impact of this decision will be discussed in Section~\ref{sec:ablation} along with remarks regarding training models of the same or better quality with a lower computational budget in an orderly fashion. In contrast, the following narrative aims at a faithful reflection of the actual way the model which we used was trained.

\paragraph{Recipe} Take\noqa{grammar-MASS_AGREEMENT} one pretrained RoBERTa$_\textsc{ large}$ model, add CRF layer and train on original (gold) dataset until progress is no longer achieved with \noqa{spell-Viterbi}Viterbi loss, SGD optimizer, and hyperparameters defined in Table~\ref{tab:hp-finetune}. Use the best-performing model to annotate random \noqa{spell-500k}500k \noqa{spell-OpenWebText}OpenWebText\footnote{See: \url{https://github.com/jcpeterson/openwebtext} OpenWebText is a project aimed at the reconstruction of OpenAI's unreleased WebText dataset.} sentences automatically. Train the second model on both original (gold) dataset and \noqa{spell-autotagged}autotagged (silver) one with hyperparameters defined in Table~\ref{tab:hp-finetune}. Repeat the procedure two more times with the best model from the previous step, hyperparameters from Table~\ref{tab:hp-finetune-self}, and other OpenWebText sentences.

Note that hyperparameters were indeed not overwritten during the first self-training iteration. Scores achieved by the best-performing models were respectively $50.91$ (without self-training) and $50.98$, $51.45$, $52.24$ in consecutive self-training iterations.

Many questions may arise regarding this procedure and the role of purely random factors. It is not a problem when rather the best score than its explanation is desired. In a leaderboard-driven exploration, one can simply conduct a broad set of experiments and choose the best-performing model without reflection, whether it is a byproduct of training instability. What actually happened here was investigated afterward and will be discussed in Section~\ref{sec:ablation}.

\begin{table} 
    \small
    \begin{minipage}{\linewidth}
        \begin{minipage}[]{0.49\linewidth}
            \input{inputs/scheme}
        \end{minipage}%
        \hfill
        \begin{minipage}[]{0.49\linewidth}
            \centering
            \input{inputs/hp-finetune}
            \vspace{5mm}
            \input{inputs/hp-finetune-self}
        \end{minipage}%
    \end{minipage}%
\end{table}

\subsection{Technique Classification}

\begin{figure}
    \centering
    \includegraphics[width=\linewidth]{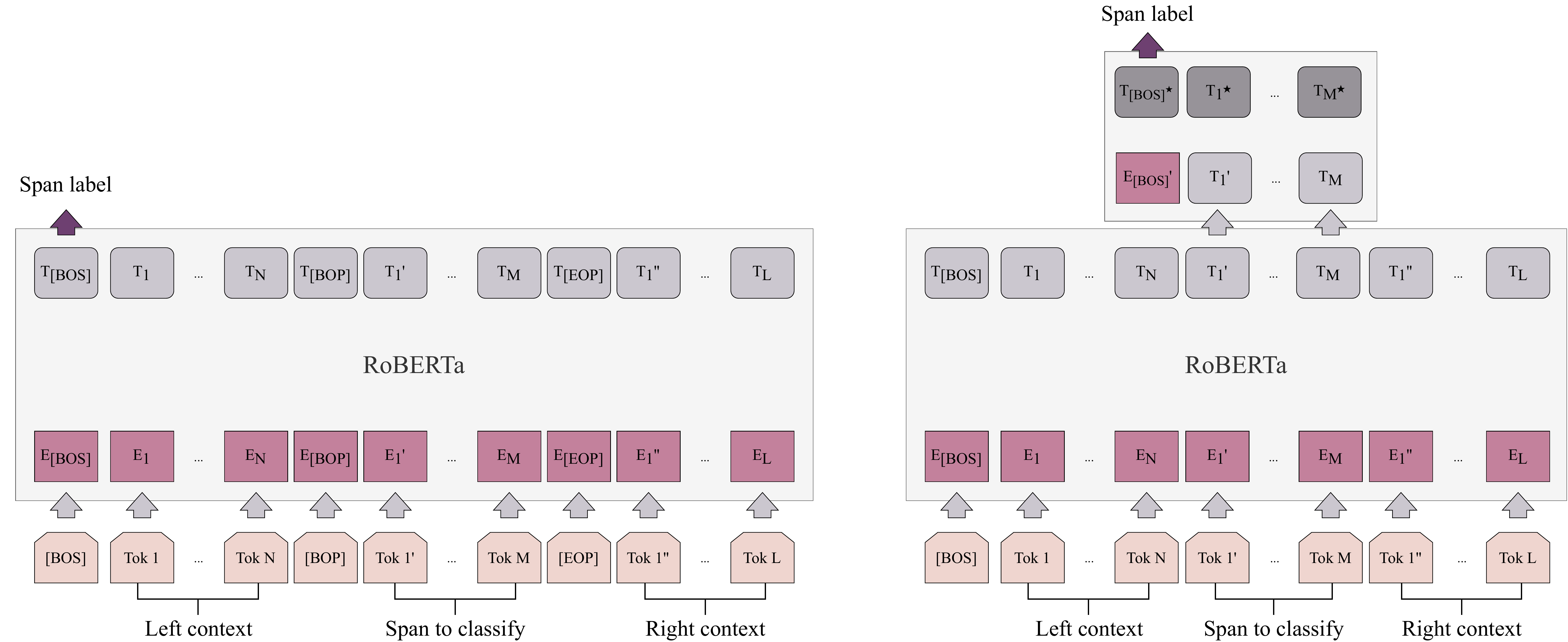}
    \caption{Comparison of span classification by means of special tokens (left) and in Span CLS approach (right). On the left, special \code{[BOP]} and \code{[EOP]} tokens are introduced, and the span is further classified as in the usual Transformer-based sentence classification task. On the right, an additional, small Transformer is stacked only over the selected tokens. It has no own embeddings apart from one for the \code{[BOS]} token, but uses representations provided by the host model instead.\label{fig:span_combined}}

\end{figure}

Transformer-based language models used in the sentence classification setting assume that representations of special tokens (such as \code{[CLS]} or \code{[BOS]}) are passed to the classification layer. Since TC task is aimed at the classification of spans, it might be beneficial to introduce information about the text fragment to be classified. We experimented with two approaches addressing this requirement.

The first assumes an injection of special tokens indicating the beginning and the end of the text marked as propaganda, such as a sample sentence before BPE applied appears as: \begin{displayquote}
    \centering
    \code{[BOS]} Democrats acted like \code{[BOP]} babies \code{[EOP]} at the SOTU \code{[EOS]}
\end{displayquote}

\noindent In this approach we continue with representation of \code{[BOS]}, as in the usual sentence classification task. The second approach is to stack a small Transformer only on the selected tokens.\footnote{The Transformer we used in our experiment had $3$ hidden layers, $4$ attention heads and an intermediate layer of size $512$. Note that hidden size depends on host model, since we are using external embeddings.} This one has no own embeddings apart from the ones for \code{[BOS]} but uses the host model's representations instead. This technique is roughly equivalent to adding consecutive layers and masking attention outside the selected span and will be referred to as Span CLS. Figure~\ref{fig:span_combined} summarizes differences between Span CLS and classification using special \code{[BOP]} and \code{[EOP]} tokens.

The initial experiments have shown that underrepresented classes achieve lower scores. To overcome this problem, we experimented with class-dependent rescaling applied to binary cross-entropy. In this setting (further referred to as \textit{re-weighting}) factor for each class was determined as its inverse frequency multiplied by the frequency of the most popular class.
The modified loss is equal to:
\begin{align*} \begin{aligned} \ell(\mathbf{x}, \mathbf{y}) & = -\frac{1}{Nd}\sum_{n=1}^N \sum_{k=1}^d\big[p^k y_n^k \log x_n^k + (1 - y_n^k)\log(1 - x_n^k) \big] \\ p^k & = \frac{1}{f^k}\max(\mathbf{f})\\ \end{aligned} \end{align*}

\noindent where $N$ is the batch size, $n$ index denotes nth batch element, $d$ is the number of classes, $\mathbf{f}$ stands for a vector of class absolute frequencies calculated on the train set, $\mathbf{x}$ is the output vector from the last sigmoid layer and $\mathbf{y}$ is a vector of multi-hot encoded ground truth labels. Note that the only difference from the original binary cross entropy for multi-label classification is the addition of the $p^k$ class weights.

In addition to the above, a part of the tested models took the use of the self-training approach. In the case of TC task one had to identify spans first and then predict their classes to generate silver train set (Figure~\ref{fig:self}). We reused our best-performing model from SI task to identify spans, and the TC model trained on ground truth to automatically annotate these spans.

Regardless of the approach taken, context as broad as possible within the 256 subword units limit was provided on both sides of the span to be classified. Note that it was a maximum equal extension of the span text in both directions, and we did not limit the extension to the sentence boundaries.

The winning TC model (described in the recipe below) was an ensemble of three models. Each of them used a different mix of previously described approaches with hyperparameters defined in Table~\ref{tab:hp-finetune} for first and second model, and those from Table~\ref{tab:hp-finetune-self} in case of the third model.
\paragraph{Recipe} Add classification layer (described in Figure~\ref{fig:span_combined} on the left) to the pretrained RoBERTa$_\textsc{ large}$ model in order to obtain the first model and train until no score gain is observed on development set. Train the second model in the same manner, but this time using the \textit{re-weighting}. Combine \textit{re-weighting}, Span CLS and self-training approaches to get the third model, and again train until no score improvement on development set is observed. Finally, ensemble all three models by averaging class probabilities from their final layers.

As shown later, the approach we took and reported above turned out to be sub-optimal. An in-depth analysis of this system and a better one is proposed in Section~\ref{sec:tc}.

\section{Ablation Studies\label{sec:ablation}}

Since different random initialization or data order can result in considerably higher scores,\footnote{See e.g.,~\citet{junczys-dowmunt-etal-2018-approaching} or recent analysis of~\citet{dodge2020finetuning}.} models with different random seeds were trained for the purposes of ablation studies. In the case of the SI task, results were evaluated on the original development set. In contrast, in the case of TC, where fewer data points are available, we decided to use cross-validation instead.

\subsection{Span Identification}

\begin{table} 
    \small
    \begin{minipage}{\linewidth}
        \begin{minipage}[b]{0.49\linewidth}
            \input{inputs/crf}
        \end{minipage}%
        \hfill
        \begin{minipage}[b]{0.49\linewidth}
            \centering
            \input{inputs/scores-si}
            \vspace{5mm}
            \input{inputs/scores-si-bad}
        \end{minipage}%
    \end{minipage}%
\end{table}

Models with different random seeds were trained for \noqa{spell-60K}60K steps with an evaluation performed every 2K steps. This is equivalent to approximately 30 epochs, and per-epoch validation in a scenario without data generated during the self-training procedure. Table~\ref{tab:scores-si} summarizes the best scores achieved across 10 runs for each configuration.

CRF has a noticeable positive impact on \noqa{spell-FLC}FLC-F1~\cite{DaSanMartinoSemeval20task11} scores achieved without self-training in the setting we consider. The presence of the CRF layer is correlated positively with the score ($\rho = 0.27$, $p < 0.001$). The difference is significant ($p < 0.001$), according to the Kruskal–Wallis test~\cite{kruskal1952use} . Unless said otherwise, all further statistical statements within this section were confirmed with statistically significant positive Spearman rank correlation and Kruskal-Wallis test results. Differences in variance were confirmed using Bartlett's test~\cite{bartlett}. The $0.05$ significance level was assumed.

The statistically significant influence of CRF disappears when the self-training is investigated. In the case of first self-training, regardless of whether or not CRF was used, a considerable increase in median score can be observed. Self-trained models with and without the CRF layer, however, are indistinguishable.

Improvement offered by further self-training iterations is not so evident but is statistically significant. In particular, they slightly improve mean scores and decrease variance (see Table~\ref{fig:crf}). As it comes to the latter, CRF-extended models generally have higher variance and scores achieved across the runs.

Table~\ref{tab:scores-si-bad} analyzes the importance of using different hyperparameters. Whereas use of a smaller batch size and dropout is beneficial for the initial training without noisy data, it negatively impacts the self-training phase. The most substantial negative impact is observed when dropout is disabled during training on the small amount of manually annotated data.

Figure~\ref{fig:crf} illustrates scores achieved by models trained for the same number of steps on subsets or supersets of manually annotated data. CRF layer has a positive impact regardless of the percentage of train set available. Once again, a large variance in scores of CRF-equipped models can be observed, however, it is substantially reduced with the increase of a batch size. Interestingly, figures suggest the proportion of automatically annotated data we used might be suboptimal since it was an equivalent of around 3000\% in line with the chart's convention. One may hypothesize better scores would be achieved by models trained with $1:4$ gold to silver proportion. 


\subsection{Technique Classification\label{sec:tc}}
6-fold cross-validation was conducted. The
results are presented in Table~\ref{tab:scores-tc}.
Folds were created by mixing training and development datasets, then shuffling them and splitting into even folds. Parameters were set according to Table~\ref{tab:hp-finetune} and Table~\ref{tab:hp-finetune-self}, whereas experiments were carried out as follows. Each approach from Table~\ref{tab:scores-tc} was separately evaluated on each fold using the micro-averaged F1 metric. Then, for each approach, the average score and the standard deviation were obtained using six scores from every fold.

\begin{table} 
    \small
    \begin{minipage}{\linewidth}
        \begin{minipage}[t]{0.49\linewidth}
            \input{inputs/scores-tc}
        \end{minipage}%
        \hfill
        \begin{minipage}[t]{0.49\linewidth}
            \centering
            \input{inputs/scores-tc-ensemble}
        \end{minipage}%
    \end{minipage}%
\end{table}

Moreover, all the 247 possible ensembles\footnote{It is the number of all subsets with cardinality greater than one, drawn from an 8-element set.} were evaluated in the same fashion as in experiments from Table~\ref{tab:scores-tc}.
Table~\ref{tab:scores-tc-ensemble} shows the performance achieved by selected combinations when simple averaging of the probabilities returned by individual models was used as the final prediction.

Due to a large number of available results, it is beneficial to conduct a statistical analysis to formulate remarks regarding the general observed trends. Each component model of the ensemble was treated as a categorical variable with respect to the ensemble score. 
Spearman rank correlation between the presence of an ensemble component (approaches from Table~\ref{tab:scores-tc}) and achieved scores shows that adding model to the ensemble correlates with a significant increase in score, except for (6) model (see Table~\ref{tab:spearman-srh}).
Boxplots\noqa{spell-Boxplots} from Figure~\ref{fig:boxplots-tc} lead to the same conclusions.\footnote{\noqa{spell-Kruskal}Kruskal-Wallis test and \noqa{spell-Boruta}Boruta algorithm~\cite{kursa2010boruta} were used in addition to support these findings too.}

Re-weighting seems to be beneficial only when ensembled with other models. An interesting finding is that Span CLS offers a small but consistent increase of performance both in models from Table~\ref{tab:scores-tc} and when used in ensembles. Bear in mind, we outperformed the second-placed team by $\varepsilon$, so an improvement of a point or half is not negligible.

What is most conspicuous, however, is that self-training based solutions from Table~\ref{tab:scores-tc} seem to be detrimental in the case of TC task. This damaging effect can be potentially attributed to the fact that automatically generated data accumulate errors from both Span Identification and Classification. Another possible explanation is that much fewer data points are available for span classification task than for span identification attempted as a sequence labeling task. The latter would be somehow consistent with what was found in the field of Neural Machine Translation, where the use of the back-translation technique in low-resource setting was determined to be harmful~\cite{Edunov2018UnderstandingBA}.

On the other hand, self-training has a positive, statistically significant impact on the score when used in ensembles (see Figure~\ref{fig:boxplots-tc} and Table~\ref{tab:spearman-srh}). It is not surprising as the beneficial impact of combining individual estimates was observed in many disciplines and is known since the times of Laplace \cite{CLEMEN1989559}.


\begin{table} 
    \small
    \begin{minipage}{\linewidth}
        \begin{minipage}[]{0.48\linewidth}
            \centering
            \input{inputs/boxplots-tc}
        \end{minipage}%
        \hfill
        \begin{minipage}[]{0.48\linewidth}
            \centering
            \input{inputs/spearman-srh}
        \end{minipage}%
    \end{minipage}%
\end{table}

\section{Error analysis}

In addition to providing an overview of problematic classes, the question of which shallow features influence score and worsen the results was addressed. This problem was analyzed \noqa{grammar-IN_A_X_MANNER}in a \textit{no-box} manner, as proposed by~\citet{gralinski-etal-2019-geval}. The main idea is to create two dataset subsets for each feature considered (one for data points with the feature present and one for data points without the feature), rank subsets by per-item scores, and use Mann-Whitney rank \textit{U}~\cite{mann1947test} to determine whether there is a non-accidental difference between subsets. A low p-value indicates that feature reduces the evaluation score of the model.

\subsection{Span Identification}

Since the \noqa{spell-FLC}FLC-F1 metric used in the SI task gives non-zero scores for partial matches; it is interesting to analyze what was the proportion of entirely missed (partially identified) spans. Table~\ref{tab:si-errors} investigates this question broken down by the propaganda technique used.

\begin{table} 
\setlength{\tabcolsep}{2mm}
\small
\centering
\begin{tabular}{lrrrrrrrrrrrrrrrr}
      & & \rotatebox{90}{Authority} & \rotatebox{90}{Fear} & \rotatebox{90}{Bandwagon} & \rotatebox{90}{B\&W} & \rotatebox{90}{Simplification} & \rotatebox{90}{Doubt} & \rotatebox{90}{Minimization} & \rotatebox{90}{Flag-Waving} & \rotatebox{90}{Loaded} & \rotatebox{90}{Labeling} & \rotatebox{90}{Repetition} & \rotatebox{90}{Slogans} & \rotatebox{90}{Clichés} & \rotatebox{90}{Strawman} & Overall \\
    \toprule
    Identified subsequence & & $57$ & $56$ & $20$ & $36$ & $50$ & $42$ & $48$ & $40$ & $44$ & $45$ & $26$ & $62$ & $41$ & $41$ & 43 \\
    Fully identified & \% & $7$ & $18$ & $0$ & $18$ & $5$ & $6$ & $11$ & $50$ & $25$ & $21$ & $33$ & $7$ & $23$ & $10$ & 23 \\
    Not identified & &  $35$ & $25$ & $80$ & $45$ & $44$ & $51$ & $39$ & $9$ & $29$ & $33$ & $40$ & $30$ & $35$ & $48$ & 33 \\
    \midrule
    Number of instances & & $14$ & $44$ & $5$ & $22$ & $18$ & $66$ & $68$ & $87$ & $325$ & $183$ & $145$ & $40$ & $17$ & $29$ & 1063 \\
    \bottomrule
\end{tabular}
\caption{Proportion of partially and fully identified spans (SI task) depending on the propaganda technique used. All the experiments conducted on the original development set.\label{tab:si-errors}}
\end{table}

Our system was unable to identify one-third of expected spans, whereas a majority of those correctly identified were the partial matches. The spans the easiest to identify in the text represented \textit{Flag-Waving}, \textit{Appeal to fear/prejudice}, and \textit{Slogans} techniques. In contrast, \textit{Bandwagon}, \textit{Doubt}, and the group \noqa{spell-Whataboutism}of \{\textit{Whataboutism}, \textit{Strawman}, \textit{Red Herring}\}  turned\noqa{proselint-cliches.write_good} out to be the hardest. The highest proportion of fully identified spans was achieved for \textit{Flag-Waving}, \textit{Repetition}, and \textit{Loaded Language}. Unfortunately, it is not possible to investigate precision in this manner, without training separate models for each label or estimating one-to-one alignments between output and expected spans.

Further investigation of problematic cases in a paradigm of no-box debugging with the GEval tool~\cite{gralinski-etal-2019-geval} revealed the most worsening features, that are features whose presence impacts span identification evaluation metrics negatively (Table~\ref{tab:geval-si}). It seems that our system tends to return ranges without adjacent punctuation. This is the case of sentences such as \textit{The new CIA Director \noqa{spell-Haspel}Haspel, who ‘tortured some folks,’ probably can’t travel to the EU}, where only the quoted text was returned, whereas annotation assumes it should be returned with apostrophes and commas. This remark can be used to improve overall results with simple post-processing slightly. Returned \textit{and} conjunction refers to the cases where it connects two propaganda spans. The system frequently returns them as a single span, contrary to what is expected in the gold standard.

\subsection{Technique Classification}

\begin{table} 
    \small
    \begin{minipage}{\linewidth}
        \begin{minipage}[b]{0.49\linewidth}
            \input{inputs/cm}
        \end{minipage}%
        \hfill
        \begin{minipage}[b]{0.49\linewidth}
            \centering
            \input{inputs/geval-si}
            \vspace{5mm}
            \input{inputs/geval-tc}
        \end{minipage}%
    \end{minipage}%
\end{table}

Figure~\ref{fig:cm_final_ensemble} presents the normalized confusion matrix of the submitted system predictions. Interestingly, there are a few commonly confused pairs. \textit{Loaded Language} and \textit{Black-and-white Fallacy} were frequently misclassified as \textit{Appeal to fear/prejudice}. Similarly, \textit{Causal Oversimplification} was often predicted as \textit{Doubt} and \textit{Clichés} as \textit{Loaded Language}.

The most worsening features are presented in Table~\ref{tab:geval-tc}. One of the frequent predictors of low accuracy is a comma character present within the span to be classified. It can  probably be attributed to the fact that its presence is a good indicator of span linguistic complexity. Another determinant of inefficiency turned out to be a negation—around half of the sentences containing word \textit{not} were misclassified by the system. Suggested features of a quotation mark before the span and the digram \textit{according to} after the span are related to reported or indirect speech. The explanation of the worsening effect of other features is not as evident as in the case mentioned above. Moreover, it seems there is no obvious way of improving the final results with our findings, and a more detailed analysis might be required.

\section{Discussion and Summary}

The winning system for the propaganda Technique Classification (TC) task and the second-placed system for the propaganda Span Identification (SI) task has been described. Both of the developed solutions used a semi-supervised learning technique of self-training. Although CRF is barely used with Transformer-based language models, the SI task was approached with RoBERTa-CRF architecture. An ensemble of RoBERTa-based models has been proposed for the TC task, with one of them making use of Span CLS layers we introduce in the present paper.

Analysis conducted afterward can be applied in a rather straightforward manner to further improve the scores for both SI and TC tasks. It is because some of the decisions we have made given lack of or uncertain information, during the post-hoc inquiry turned out to be sub-optimal. These include the proportion of data from self-training in the SI task, and the possibility of providing a better ensemble in the case of TC.

The ablation studies conducted, however, have some limitations. The same subset of \noqa{spell-OpenWebText}OpenWebText was used in experiments conducted within one self-training iteration. This means a random seed did not impact which sentences were used during the first, second, and third self-training phase, and in each, we were manipulating only the data order. Moreover, an analysis we reported was limited to few hyperparameter combinations and no extensive hyperparameter space search was performed. Finally, only one and a rather simple method of cost-sensitive re-weighting was tested, and there is a great chance it was sub-optimal. It would be interesting to investigate other schemes, such as the one proposed by~\citet{Cui2019ClassBalancedLB}.\noqa{grammar-TO_TOO}

The error analysis revealed propaganda techniques commonly confused in TC task, and the techniques we were unable to detect effectively within the SI input articles. In addition to providing an overview of problematic classes, the question of which shallow features influence score and worsen the results was addressed. A few of these were identified and our remarks can be used to slightly improve results on SI task with simple post-processing. This is not the case for TC task, where one is unable to propose how to improve the final results with our findings.

An interesting future research direction seems to be the application of the CRF layer and Span CLS to Transformer-based language models when dealing with other tasks outside the propaganda detection problem. These may include Named Entity Recognition in the case of RoBERTa-CRF, and an aspect-based sentiment analysis that can be viewed through the lens of span classification with Span CLS we proposed.


\section{Outro}\noqa{spell-Outro}

Developed systems were used to identify and classify spans in the present paper to detect fragments one may suspect to represent one or more propaganda techniques. Unfortunately for the entertaining value of this work, none of such were identified by our SI model.

%% file: inputs/scheme.tex
\begin{figure}[H]
    \centering
    \includegraphics[width=0.6\linewidth]{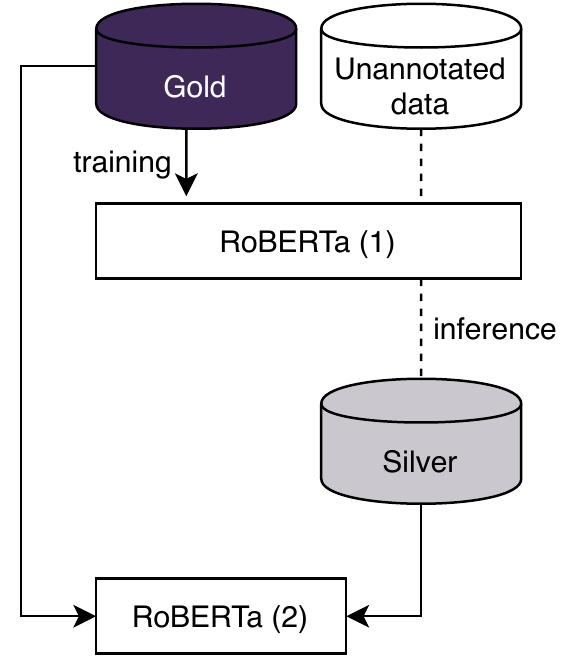}
    \caption{Self-training stands for a process of training an initial model on manually annotated dataset first and using it to further extend train by means of annotating other dataset automatically.}
    \label{fig:self}
\end{figure}

%% file: inputs/hp-finetune.tex
\begin{tabular}{lcccc}
    \toprule
    \bf Hparam & \bf SI & \bf TC \\
    \midrule
    Dropout             & \multicolumn{2}{c}{.1}    \\
    Attention dropout   & \multicolumn{2}{c}{.1}    \\
    Max sequence length & 256  & 256 \\
    Batch size          & 8    & 16 \\
    Learning rate       & 5e-4 & 2e-5 \\
    Number of steps     & 60k  & 20k \\
    Learning rate decay &
    \multicolumn{2}{c}{--} \\
    Weight decay        & -- & .01 \\
    Momentum            & .9 & -- \\
    \midrule
    Optimizer           & SGD & AdamW \\
    Loss                & Viterbi & BCE \\
    \bottomrule
\end{tabular}
\caption{
  Optimizers and hyperparameters used for both finetuning RoBERTa and training additional parameters. 
}
\label{tab:hp-finetune}

%% file: inputs/hp-finetune-self.tex
\begin{tabular}{p{36mm}cccc}
    \toprule
    \bf Hparam & \bf SI & \bf TC \\
    \midrule
    Dropout             & \multicolumn{2}{c}{.0}    \\
    Attention dropout   & \multicolumn{2}{c}{.0}    \\
    Batch size          & 16 & 16  \\
    \bottomrule
\end{tabular}
\caption{
    Hyperparameter overwrites for self-training.
}
\label{tab:hp-finetune-self}

%% file: inputs/crf.tex
\begin{figure}[H]
    \centering
    \includegraphics[width=0.9\linewidth]{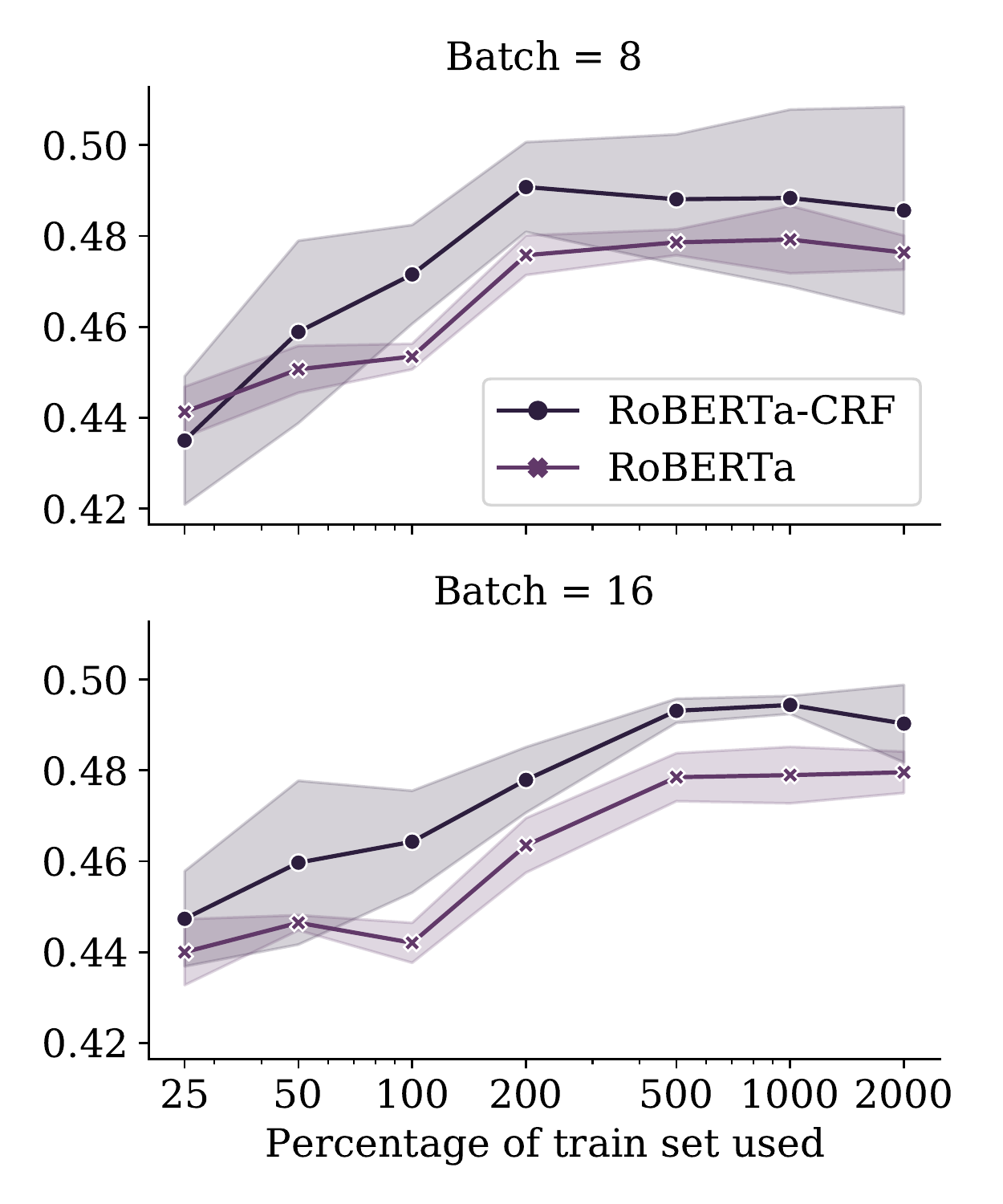}
    \caption{Performance of RoBERTa with and without CRF as a~function of percentage of train set available. Values above 100\% indicate self-training was performed. Mean FLC-F1 and standard deviation across 5 runs for each percentage.\label{fig:crf}}
\end{figure}

%% file: inputs/scores-si.tex
\begin{tabular}{ccc}
    \toprule
    \bf CRF & \bf Self-train & {\bf FLC-F1} (std, max) \\
    \midrule
    $-$ & $-$ & $45.2 \pm 0.3$ \quad $45.6$ \\
    $+$ & $-$ & $47.4 \pm 0.8$ \quad $48.2$  \\
    $-$ & $+$ & $48.9 \pm 0.5$ \quad $50.2$ \\
    $+$ & $+$ & $49.1 \pm 3.0$ \quad $51.7$ \\
    $+$ & \hspace{4.2mm} $+$ $(2)$ & $49.7 \pm 2.0$ \quad $51.6$ \\
    $+$ & \hspace{4.2mm} $+$ $(3)$ & $50.0 \pm 1.8$ \quad $51.8$ \\
    \bottomrule
\end{tabular}
\caption{Best scores on the dev set achieved with RoBERTa large model on SI task. Mean, standard deviation and maximum across 10 runs with different random seeds. Numbers in brackets indicate how many self-training iterations were used.}
\label{tab:scores-si}

%% file: inputs/scores-si-bad.tex
\begin{tabular}{ccccc}
    \toprule
    \bf Batch & \bf Dropouts & \bf Self-train
    & \bf CRF
    & $\Delta$ {\bf FLC-F1} \\
    \midrule
    \multirow{4}{*}{$16 \rightarrow 8$} &
    \multirow{2}{*}{$.0 \rightarrow .1$} &
    \multirow{4}{*}{$+$} &
    $-$ &
    $-1.1$ \\
    &
    &
    &
    $+$ &
    $-1.6$ \\
    &
    \multirow{2}{*}{$.0$} &
    &
    $-$ &
    $-0.4$ \\
    &
    &
    &
    $+$ &
    $-1.1$ \\
    \midrule
    \multirow{4}{*}{$8 \rightarrow 16$} &
    \multirow{2}{*}{$.1 \rightarrow .0$} &
    \multirow{4}{*}{$-$} &
    $-$ &
    $-3.9$ \\
    &
    &
    &
    $+$ &
    $-7.0$ \\
    &
    \multirow{2}{*}{$.1$}&
    &
    $-$ &
    $-0.7$ \\
    &
    &
    &
    $+$ &
    $-1.3$ \\
    \bottomrule
\end{tabular}
\caption{Impact of hypothetical lowering batch size during self training or enlarging batch size during initial training, as well as of enabling or disabling both hidden and attention dropouts. Change between means across 10 runs with different random seeds.}
\label{tab:scores-si-bad}

%% file: inputs/scores-tc.tex
\setlength{\tabcolsep}{1.5mm}
\begin{tabular}{ccccc}
    \toprule
    \bf \# & \bf Re-weight & \bf Span CLS & \bf Self-train & {\bf Micro-F1} (std) \\
    \midrule
    (1) & $-$ & $-$ & $-$ & $71.9 \pm 1.5$  \\
    (2) & $-$ & $-$ & $+$ & $71.4 \pm 1.4$  \\
    (3) & $-$ & $+$ & $-$ & $72.2 \pm 1.3$  \\
    (4) & $-$ & $+$ & $+$ & $71.8 \pm 1.7$  \\
    (5) & $+$ & $-$ & $-$ & $71.8 \pm 1.6$  \\
    (6) & $+$ & $-$ & $+$ & $70.9 \pm 1.7$  \\
    (7) & $+$ & $+$ & $-$ & $72.4 \pm 1.5$  \\
    (8) & $+$ & $+$ & $+$ & $71.3 \pm 1.5$  \\
    \bottomrule
\end{tabular}
\caption{Average of 6-fold cross-validation score on TC task with micro-averaged F1 metric.}
\label{tab:scores-tc}

%% file: inputs/scores-tc-ensemble.tex
\begin{tabular}{lc}
\toprule
\bf Ensemble \bf & {\bf Micro-F1} (std) \\
\midrule
(1) (6) & $72.3 \pm 1.7$ \\
(1) (2) & $72.9 \pm 1.8$ \\
(3) (5) & $73.6 \pm 1.5$ \\
(1) (5) (8) & $74.1 \pm 1.7$ \\
(2) (4) (7) & $74.4 \pm 1.5$ \\
(1) (4) (7) & $74.6 \pm 1.4$ \\
(1) (4) (7) (8) & $74.9 \pm 1.2$ \\
(1) (2) (4) (5) (7) & $75.1 \pm 1.5$ \\
\bottomrule
\end{tabular}
\caption{Average scores achieved with ensembles of individual models described in Table~\ref{tab:scores-tc}. Micro-averaged F1 metric.\label{tab:scores-tc-ensemble}}

%% file: inputs/boxplots-tc.tex
\begin{figure}[H]
    \centering
    \includegraphics[width=\linewidth,trim={0.5cm 0 0.4cm 0},clip]{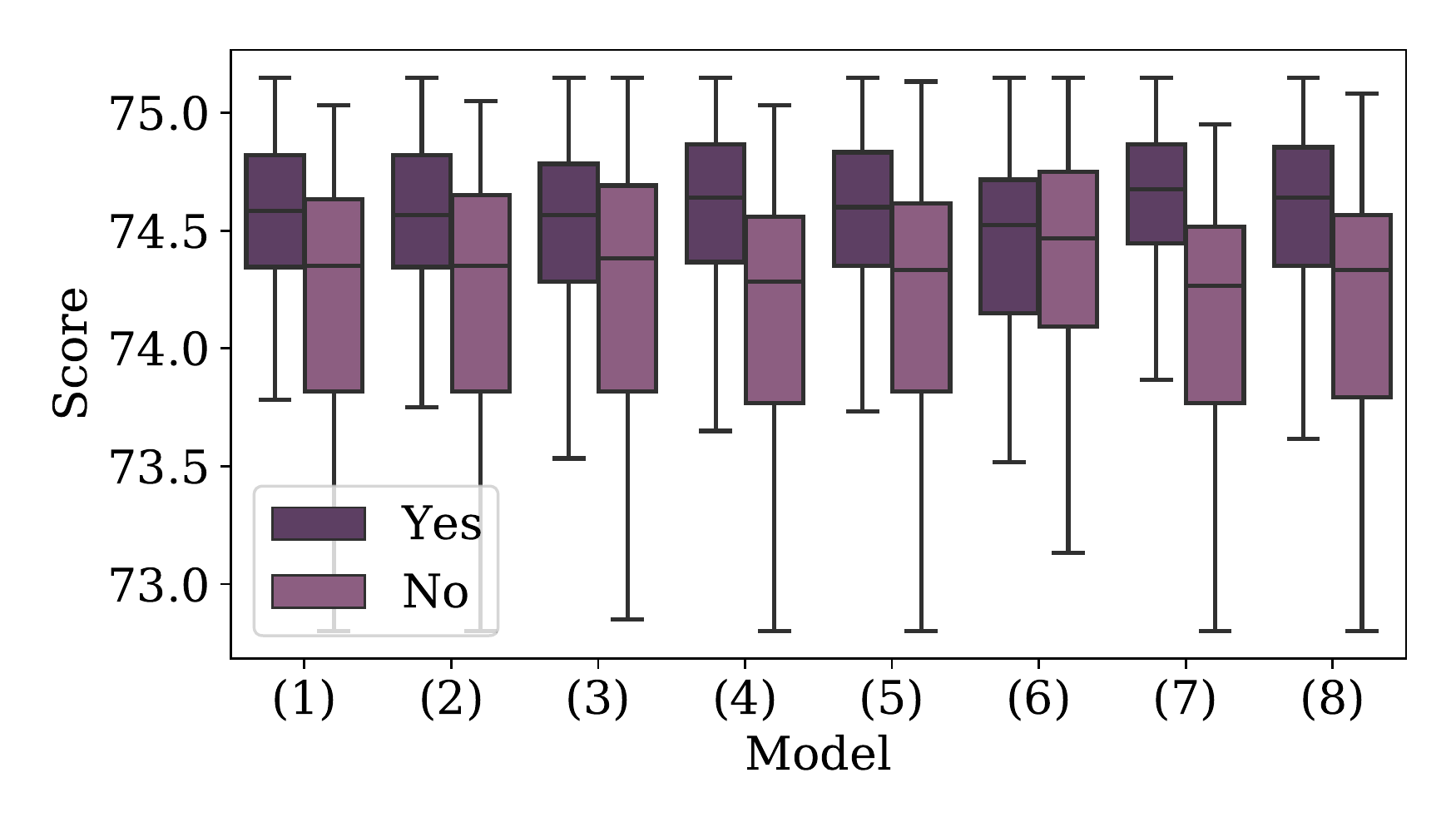}\vspace{-3mm}
    \caption{Impact of adding a particular model to the ensemble has on mean scores from different folds. Comparison of results with and without it present in tested combination.}     
    \label{fig:boxplots-tc}
\end{figure}{}

%% file: inputs/spearman-srh.tex
\setlength{\tabcolsep}{1.9mm}

    \begin{tabular}{crrrrrrrr}
    \toprule
    \bf Model & (1) & (2) & (3) & (4) & (5) & (6) & (7) & (8) \\
    \midrule
    $\bm{\rho}$ & $.28$ & $.30$ & $.20$ & $.41$ & $.32$ & $.05^*$ & $.50$ & $.36$ \\
    \bottomrule
    \end{tabular}
    \caption{Spearman's $\rho$ between presence of ensemble component (models from Table~\ref{tab:scores-tc}) and score achieved by ensemble. $^*$ indicate results were not significant, assuming $0.05$ significance level.}
    \label{tab:spearman-srh}

%% file: inputs/cm.tex
\begin{figure}[H]
    \centering
    \includegraphics[width=\linewidth,trim={1.5cm 0 3cm 0},clip]{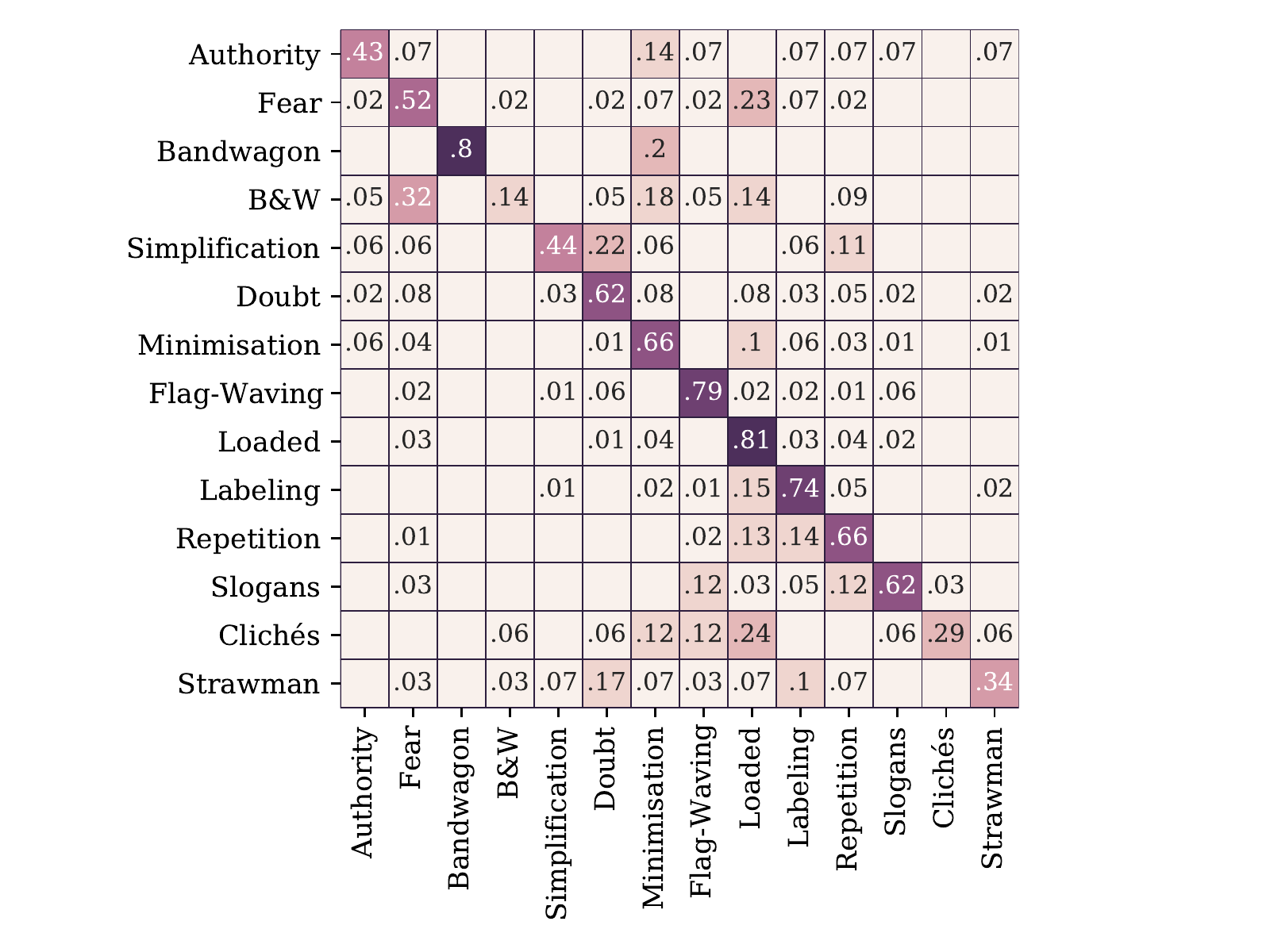}
    \caption{Confusion matrix of the submitted system predictions normalized over the number of correct labels. Rows represent the correct labels and columns -- the predicted ones (TC).}
    \label{fig:cm_final_ensemble}
\end{figure}

%% file: inputs/geval-si.tex
\begin{tabular}{llrrr}
    \toprule
    \bf Feature & & \bf Count & \bf P-value \ \\
    \midrule
    \textit{question} & expected & $21$ & $0.036$ \\
    \textit{dot} & & $36$ & $0.037$ \\
    \textit{quotation} & & $58$ & $0.050$ \\
    \textit{exclamation} & & $15$ & $0.064$ \\
    \midrule
    and & output & $14$ & $0.070$ \\
    \bottomrule
\end{tabular}
\caption{Selected shallow features one may hypothesize impact evaluation scores negatively (SI).}
\label{tab:geval-si}

%% file: inputs/geval-tc.tex
\begin{tabular}{llrrr}
    \toprule
    \bf Feature & & \bf Count & \bf P-value \ \\
    \midrule
    \textit{comma} & inside & $119$ & $<0.001$ \\
    we & & $15$ & $0.002$ \\
    this & & $28$ & $0.007$ \\
    will & & $40$ & $0.008$ \\
    not & & $62$ & $0.013$ \\
    \textit{exclamation} &  & $16$ & $0.014$ \\
    \midrule
    CIA & before & $25$ & $<0.001$ \\
    according to & after & $8$ & $<0.001$ \\
    \textit{quotation} & before & $65$ & $0.004$ \\
    \bottomrule
\end{tabular}
\caption{Selected shallow features one may hypothesize impact evaluation scores negatively (TC).}
\label{tab:geval-tc}